\documentclass[journal]{IEEEtai}

\usepackage[colorlinks,urlcolor=blue,linkcolor=blue,citecolor=blue]{hyperref}

\usepackage{color,array}
\usepackage{amsmath}
\usepackage{amssymb}
\usepackage{graphicx}
\usepackage{booktabs}
\usepackage{multirow}
\usepackage{subcaption}
\usepackage{orcidlink}
\usepackage{algorithm}
\usepackage{algorithmic}
\usepackage{balance}

\begin{document}

\title{FairCompass: Operationalising Fairness in Machine Learning} 

\author{Jessica Liu, Huaming Chen\orcidlink{0000-0001-5678-472X}~\IEEEmembership{Member,~IEEE}, Jun Shen\orcidlink{https://orcid.org/0000-0002-9403-7140}~\IEEEmembership{Senior Member,~IEEE}, and Kim-Kwang Raymond Choo\orcidlink{0000-0001-9208-5336}
\thanks{Jessica Liu and Huaming Chen are with the School of Electrical and Computer Engineering, University of Sydney, Australia. (Corresponding author e-mail: huaming.chen@sydney.edu.au). Jun Shen is with the University of Wollongong, Australia. (e-mail: jshen@uow.edu.au). Kim-Kwang Raymond Choo is with The University of Texas at San Antonio, San Antonio, TX 78249, USA. (e-mail: raymond.choo@fulbrightmail.org).}
}

\markboth{}
{}

\maketitle

\begin{abstract}
As artificial intelligence (AI) increasingly becomes an integral part of our societal and individual activities, there is a growing imperative to develop responsible AI solutions. Despite a diverse assortment of machine learning fairness solutions is proposed in the literature, there is reportedly a lack of practical implementation of these tools in real-world applications. Industry experts have participated in thorough discussions on the challenges associated with operationalising fairness in the development of machine learning-empowered solutions, in which a shift toward human-centred approaches is promptly advocated to mitigate the limitations of existing techniques. In this work, we propose a human-in-the-loop approach for fairness auditing, presenting a mixed visual analytical system (hereafter referred to as `FairCompass'), which integrates both subgroup discovery technique and the decision tree-based schema for end users. Moreover, we innovatively integrate an Exploration, Guidance and Informed Analysis loop, to facilitate the use of the Knowledge Generation Model for Visual Analytics in FairCompass. We evaluate the effectiveness of FairCompass for fairness auditing in a real-world scenario, and the findings demonstrate the system's potential for real-world deployability. We anticipate this work will address the current gaps in research for fairness and facilitate the operationalisation of fairness in machine learning systems. 

\end{abstract}

\begin{IEEEImpStatement}
To address the drawbacks of existing fairness solutions and assist in operationalising fairness, we propose a new approach to fairness tool that combines technical, non-technical and visual analytics solutions to fairness. We demonstrate this novel approach by benchmarking with FairVis and Fairness Compass. The problem space of this paper assumes the scope of the tools, which is to assist in fairness auditing for machine learning classifiers. Ultimately, we anticipate this work as a first step towards operationalising fairness, in the direction of establishing formal fairness processes within teams, organisations, and institutions, by better meeting the needs of practitioners who are tasked with the responsibility of fairness.

\end{IEEEImpStatement}


\section{Introduction}
\IEEEPARstart{T}{he} use of artificial intelligence in decision making has become increasingly popular across a range of industries such as healthcare, commerce, marketing, education and many more~\cite{Jordan2015}. As AI takes on the automation of crucial decision-making processes traditionally overseen by humans, we observe the increasing prevalence of unfair outcomes attributed to AI systems. It is coupled with an increasing focus on fairness research and a growing call for responsible AI. Many case studies of unfair machine learning models underscore the fact that the biases within the machine learning life cycle, can adversely affect the individuals, with an overwhelming number of people belonging to marginalised groups~\cite{Mehrabi2021,Richardson2021}. 

Existing fairness research in the field of machine learning is primarily focused on addressing the challenge of minimising algorithmic bias~\cite{Richardson2021}, which has led to an abundance of technical solutions to combat algorithmic bias~\cite{Bellamy2019,Saleiro2018}. However, a growing concern revolves around the inadequate implementation of these solutions in practical settings. It highlights a disconnect between the progress in fairness detailed in literature and the practical adoption by practitioners or organisations~\cite{Holstein2019,Madaio2020,Veale2017}. One main concern of current fairness solutions is the overemphasis on the algorithmic and statistical approach to fairness, such as the statistical definitions of fairness and bias detection algorithms~\cite{Richardson2021}. It is widely recognised that fairness in machine learning is a socio-technical problem, and a human-centred approach to model understanding, diagnosing, and steering is needed to incorporate human values. This allows practitioners to pick up on information or patterns that may otherwise be neglected by computers~\cite{Sacha2017}. Furthermore, fairness practices have yet to be widely adopted by institutions and organisations as standard procedure, and the responsibility of fairness is often left to the practitioners as extra work that is high effort and low reward~\cite{Holstein2019}. Operationalising fairness in real-world problems require more nuanced and complex solutions that take into consideration the actors at each stage, refering to as human-in-the-loop machine learning. 

In~\cite{Cabrera2019}, the incorporation of fairness in machine learning with visual analytics is explored with the outcome of FairVis, an interactive system designed to audit the fairness and facilitate the discovery of biases. A main focus in FairVis is to identify the intersectional bias. More recently, the Fairness Compass is developed to formalise the decision process with a decision tree and specify the fairness objective for the application context~\cite{Ruf2021}. The decision tree serves as a guideline to narrow down the solution space, providing extensible capability to prioritise the most appropriate fairness metrics, and consequently being able to select the most suitable bias mitigation measures. Other existing works, such as Aequitas~\cite{Saleiro2018}, AI Fairness 360~\cite{Bellamy2019}, Google's What-If Tool~\cite{Wexler2020}, also present promising results in the venue of novel approaches to fairness research as they aim to address problems commonly encountered by practitioners according to fairness related literature. However, they have either overemphasis on technical solutions or difficulties in operationlising fairness for practitioners. 

To address the drawbacks of existing fairness solutions and assist in operationalising fairness, we propose `FairCompass' that combines technical, non-technical and visual analytics solutions to fairness. We demonstrate this combining the subgroup discovery technique and decision tree-based schema for end users, in which the solution is driven with human-in-the-loop. The problem assumes the scope of the tools, which is to assist in fairness auditing for machine learning classifiers. To our best knowledge, this work is a first step towards operationalising fairness, by better meeting the needs of practitioners tasked with the responsibility of fairness. 

The main contributions of this paper are:
\begin{itemize}
    \item We thoroughly review the existing AI fairness solutions, providing the context to investigate the common issues presented in the literature.
    \item We develop `FairCompass', a novel mixed visual analytics system with subgroup discovery technique and decision tree-based schema for fairness auditing.
    \item The Exploration, Guidance and Informed Analysis Loop is leveraged to facilitate the use of knowledge Generation Model for Visual Analytics in FairCompass.
    \item We evaluate FairCompass with a real-world scenario for fairness auditing to demonstrate the effectiveness of the system. The replication package is released publicly\footnote{https://github.com/Huaming-Chen/FairCompass}.
\end{itemize}

The rest of this paper starts with an overview of the state-of-the-art research in AI fairness in Section 2. Section 3 discusses the design challenges and our overall approach design process. Section 4 presents the design and implementation of FairCompass, and our conceptual framework that supports fairness auditing. We evaluate FairCompass in Section 5 with real-world case analysis. Section 6 covers the limitations and future work, and we conclude the paper in Section 7.

\section{Background}
\subsection{Unfairness in AI Systems}
The neutrality fallacy is a concept introduced as the misconception that machine learning systems are impartial to human biases~\cite{Sandvig2014}. This fallacy can be found in many AI systems over the past few years. One example is the COMPAS (Correctional Offender Management Profiling for Alternative Sanctions), which is a machine learning software used by U.S. courts to measure the likelihood of recidivism of defendants. An investigation found that COMPAS was biased against African American offenders, disproportionally predicting them as having a higher risk of recidivism, with a false positive rate of nearly double their Caucasian counterparts~\cite{Angwin2016}. Also, it is found that COMPAS model did not outperform non-expert human judgement in terms of accuracy or fairness~\cite{Dressel2018}. Similarly, ~\cite{Harrison2020} suggested that the perceived fairness of realistic machine learning models is overestimated, as their study on COMPAS did not demonstrate that a machine learning model was significantly more accurate than human judgement. 

Some popular machine learning-empowered systems can refer to as online job ad recommendation systems.~\cite{Lambrecht2019} presents an analysis of data from a field test for a STEM (Science, Technology, Engineering and Math) job ad. The ad was designed to promote job opportunities and training in STEM, with an explicit intent to deliver the ad in a fair, gender-neutral manner. However, the ad was shown empirically to over 20\% more men than women. The disparity was not due to the common perception of women being less likely to click on STEM job ads, rather suggestive evidence pointed to women being more expensive to show ads to compared to men. 

These works have collectively suggested a misplaced over-reliance on machine learning systems, even when the real-world problems in these scenarios have proved too large and complex to be fully automated without human intervention. These case studies point out the need for practitioners to attempt to understand why discriminatory outcomes are produced, rather than viewing machine learning models as a black box. Hence, fairness in responsible AI is a socio-technical issue that must be brought to the attention of the developers of machine learning systems, as well as other actors in each stage of the machine learning lifecycle. 

\subsection{Fairness Definition}
A popular definition of fairness is presented by Mehrabi et al.~\cite{Mehrabi2021}, concerning `the absence of any prejudice or favouritism toward an individual or group based on their inherent or acquired characteristics'. Thus, most fairness literature typically approaches the problem of defining fairness with three primary ideas of fairness: individual, group and subgroup.

Individual fairness requires a machine learning model to give similar predictive outcomes to similar individuals. Dwork et al. formulate the framework of fairness through awareness, which captures fairness through the principle of classifying similar individuals (with respect to certain attributes) similarly~\cite{Dwork2012}. For instance, in a loan allocation scenario, individuals with repayment rates that are similar shall receive a similar loan. Joseph et al. develop an approach allowing to distinguish `high quality' candidates and promote meritocracy, which are commonly used in decisions related to opportunity~\cite{Joseph2016}. 

\begin{table*}[htbp]
\centering
\caption{Fairness solutions proposed in literature, categorised with the features and fairness notions that are supported.}\label{tab1}
\begin{tabular}{|c|c|c|c|c|c|c|}
\hline
Fairness tool &  Bias understanding & Bias detection & Bias mitigation & Individual fairness & Group \newline fairness & Subgroup fairness\\
\hline
Aequitas & \checkmark & \checkmark & & & \checkmark &\\
\hline
AIF360 & & \checkmark & \checkmark & \checkmark & \checkmark & \\
\hline
WIT & \checkmark & \checkmark & & & & \\
\hline
Fairlearn & & \checkmark & \checkmark & & \checkmark & \\
\hline
Fairness checklist & & & & & & \\
\hline
FairSight & \checkmark & \checkmark & \checkmark & \checkmark & \checkmark & \\
\hline
\end{tabular}
\end{table*}
Group fairness requires a machine learning model to treat different groups equally~\cite{Mehrabi2021}. This definition of fairness is the most used in the development of technical fairness solutions and bias mitigation methods in literature as they can be obtained without making any assumptions about the setting, which leads to immediately actionable algorithmic solutions~\cite{Kearns2017}. In the context of group fairness, a population is partitioned into privileged and unprivileged groups based on sensitive attributes. In an unfair AI system, privileged groups are usually given more favourable outcomes, compared to underprivileged groups who are often disadvantaged due to a range of pre-existing biases~\cite{Chen2023}. Group fairness attempts to correct these biases by treating all groups equally. 

Subgroup fairness is a relatively new form of fairness that attempts to address the shortcomings of individual and group fairness and taking into consideration the intersectionality of bias. Problems with the intersectionality of bias arise in real world problems when populations are defined by multiple features. In the Gender Shades study~\cite{Buolamwini2018}, Buolamwini \& Gebru investigate a facial recognition software and evaluate the classification accuracy of subgroups based on sex and skin colour. When observing subgroups by sex and skin colour individually, they found that classifiers performed better on male faces compared to female faces with a difference of up to 20\% in error rate, and performed better on lighter faces compared to darker faces with a difference of up to 19\% in error rate. While these differences in performance were already significant, there were more drastic disparities between intersectional subgroups with both attributes, with darker skinned females having an accuracy as low as 65\%, and lighter skinned males with an accuracy of almost 100\%.

\subsection{Existing Fairness Solutions}
A large amount of fairness research has focused on tools to help mitigate data and algorithmic bias, while a variety of toolkits have been developed with the intention of helping developers and practitioners across a large range of projects and domains. Some tools such as Google's What-If Tool~\cite{Wexler2020} and FairSight~\cite{Ahn2020}, provide features that facilitate a better understanding of possible biases and how they can impact the predictive output. Doing so lays the necessary groundwork for users to be able to identify biases and address them. Most toolkits include bias detection and auditing features such as Aequitas~\cite{Saleiro2018} and Google's What-If Tool~\cite{Wexler2020}, with a few that provide libraries of bias mitigation algorithms, i.e., Microsoft’s Fairlearn~\cite{fairlearn2020}, AI Fairness 360~\cite{Bellamy2019} and FairSight~\cite{Ahn2020}. At the same time, non-technical solutions such as Microsoft’s AI fairness checklists~\cite{Madaio2020}, have been proposed to guide practitioners to carry out the appropriate fairness practices throughout their projects. In more recent years, many experts have taken a human-centred approach to fairness and bias mitigation by introducing toolkits in the form of visual analytics systems. The solutions covered in this section are solutions that have been proposed by organisations as well as academia. Table~\ref{tab1} summarises the features and fairness notions.

While diverse in application and techniques, these solutions are often criticised for various shortcomings, with common issues in the literature including: 

1) Overemphasis on Technical Solutions. It is found that overemphasis on statistical and algorithmic approaches to fairness could have a negative effect on fairness in AI systems. These solutions often fail to address the socio-technical aspect of AI fairness, and how social systems are complex, dynamic, and adaptive~\cite{Holstein2019}. This may even result in technical bias~\cite{Friedman1996}. Therefore, it is crucial to adopt frameworks to facilitate the proper application of these tools and focus on how bias manifests throughout the entire machine learning lifecycle~\cite{Rakova2020}. 

2) Subjectivity of Fairness. 23 types of biases and 10 different fairness metrics are defined in~\cite{Mehrabi2021}, concluding that the synthesis of a unified definition of fairness is one major challenge. Due to the vast range of use cases covered by proposed fairness definitions and metrics, it is inevitable that there are disparities between them that make them incompatible with each other. It is also evident that that certain fairness definitions cannot coexist, and prioritising certain metrics over others can lead to misconceptions of the fairness~\cite{Kleinberg2016,Verma2018}.

3) Difficulties in Operationalising Fairness. Recent studies suggests that the tools often fail to address the needs of practitioners~\cite{Greene2019,Holstein2019,Madaio2020,Rakova2020}, highlighting that this is an issue with the intersection of technical and design expertise. 

4) Lack of Support for Practitioners. While AI fairness is a socio-technical problem, it often leaves practitioners feeling overwhelmed and unsupported. Some are concerned as unqualified to make decisions since lacking knowledge in fairness research~\cite{Holstein2019}, and others fear missing potential biases but lack the know-how to assess their systems for unfairness~\cite{Madaio2020}.

\section{Methodology}
This section presents the proposed approach of `FairCompass'. The overarching goal for the design is to take a first step in operationalising fairness in the industry, promote the transition of fairness procedures as a formal part of the machine learning development workflow, and encourage organisations to prioritise fairness and offer better support for practitioners. With the feedback from practitioners~\cite{Holstein2019,Madaio2020,Rakova2020,Cramer2019},  the design process is informed by the practitioner feedback gathered in this work with emphasis to supplement statistical definitions with social practices~\cite{Birhane2021,Richardson2021}. 

Since the feedback from industry practitioners has the concern of a lack of application of these tools, despite the large assortment of technical and non-technical solutions~\cite{Holstein2019,Richardson2021}. A viable solution is to design a human-in-the-loop solution combining the strengths of automated or algorithmic methods with a non-technical solution to guide human decision-making. Though a tool bundle for AI fairness is presented combining the Fairness Compass and a Fairness Library~\cite{Ruf2022}, it is insufficient in operationalising fairness. An effective solution must provide more structure in the fairness application process so that it can be adopted by practitioners, teams, and organisations. Thus, we assert that, an interactive visual analytics application will capture the human-in-the-loop approach to fairness, while also providing support for operationalising fairness. This addresses the need for practitioners to have a deeper understanding of their data by learning through interactions with visualisations and allows organisations and teams to streamline the fairness process. Furthermore, it can enable the organisations to establish formal fairness-assurance processes in the machine learning development life cycle. 

\begin{figure}
\centering
\includegraphics[width=\linewidth]{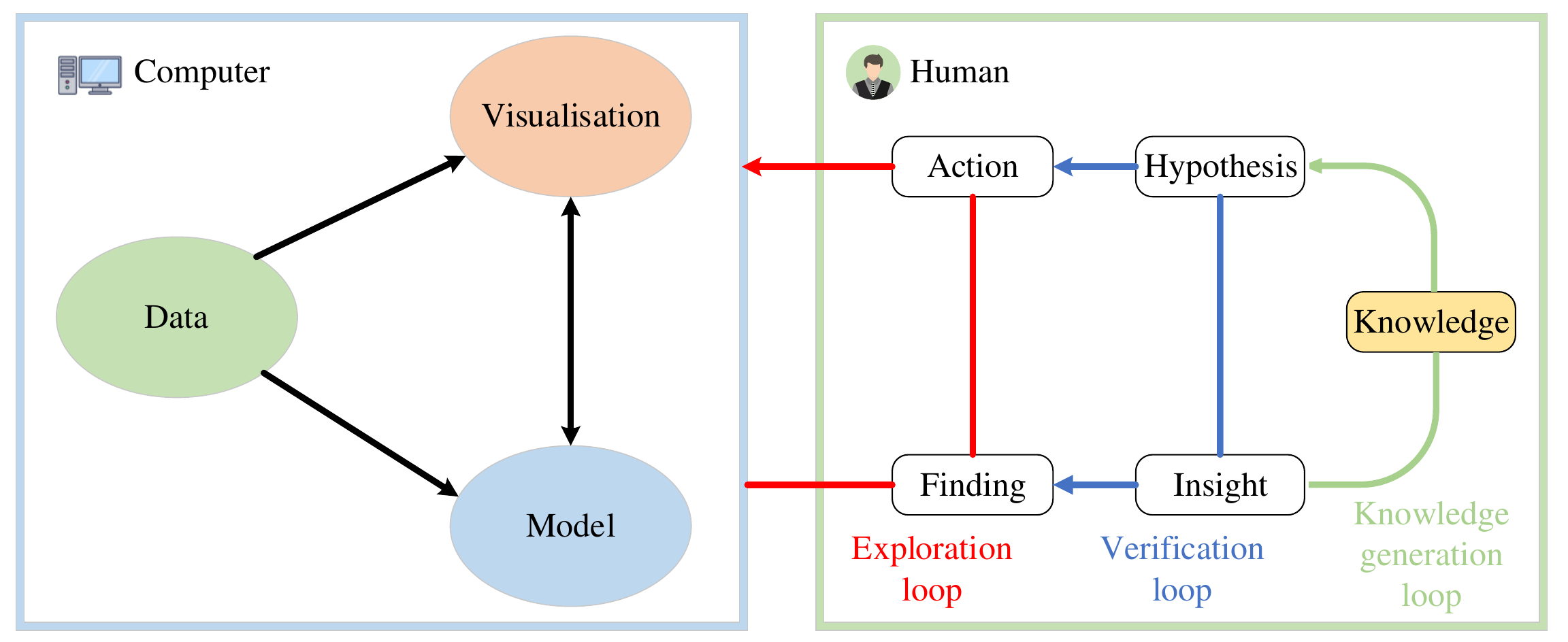}
\caption{The Knowledge Generation Model used in FairCompass for Visual Analytics~\cite{Sacha2014}} \label{fig1}
\end{figure}

Thus, we design the system as a mixed visual analytics system that allows users to view the data and apply fairness metrics, together with a decision tree-based schema that acts as a guide to select the most appropriate fairness metric. Moreover, the system supports subgroup discovery as the technical component, demonstrating the potential to be polished with visual analytics principles and models such as Sacha et al.'s Knowledge Generation Model for Visual Analytics~\cite{Sacha2014}, as shown in Fig.~\ref{fig1}. We refrain from using checklists since it may be too tedious, overwhelming, and time-consuming for practitioner workflow following Cramer et al.'s user study~\cite{Cramer2019}. Thus, as the non-technical component of the system, we leverage the decision tree-based schema as it provides a large array of fairness definitions in a clear visual representation.

As the technical and non-technical components are designed independently, we apply a conceptual framework to mesh the two solutions together. To allow for seamless integration and apply the Knowledge Generation Model for Visual Analytics, we design a new conceptual framework: the Exploration, Guidance, and Informed Analysis Loop. With the mentioned models and framework, we anticipate the proposed system properly address the design challenges, including the identified intersectional fairness, incomplete visual representation and the need for a human-centred design. Especially, we have presented the systems without specific user expectations, designed without assumptions about the user's expertise in machine learning fairness.
It has largely reduced the occurrence of the `gulfs' from Norman’s Seven Stages of Action cycle~\cite{Norman2002}, that arise when human interacts with digital interfaces. 

\begin{figure}
\centering
\includegraphics[width=\linewidth]{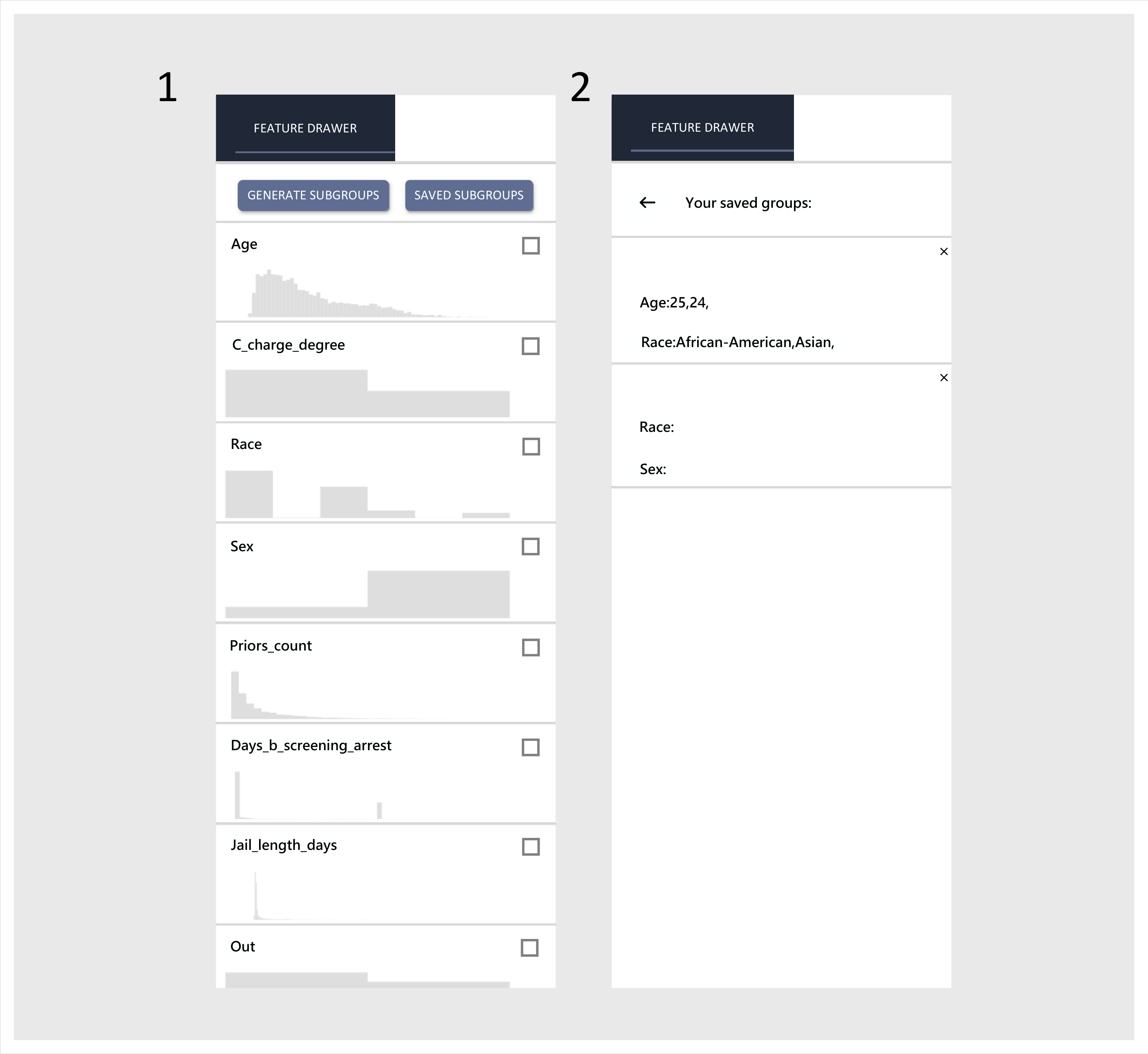}
\caption{The Feature Distribution View in FairCompass. The subgroup generation functionality (1) and the user's saved groups (2).} \label{fig2}
\end{figure}
\section{FairCompass Overview}
\begin{figure*}[htbp]
\centering
\includegraphics[width=0.75\textwidth]{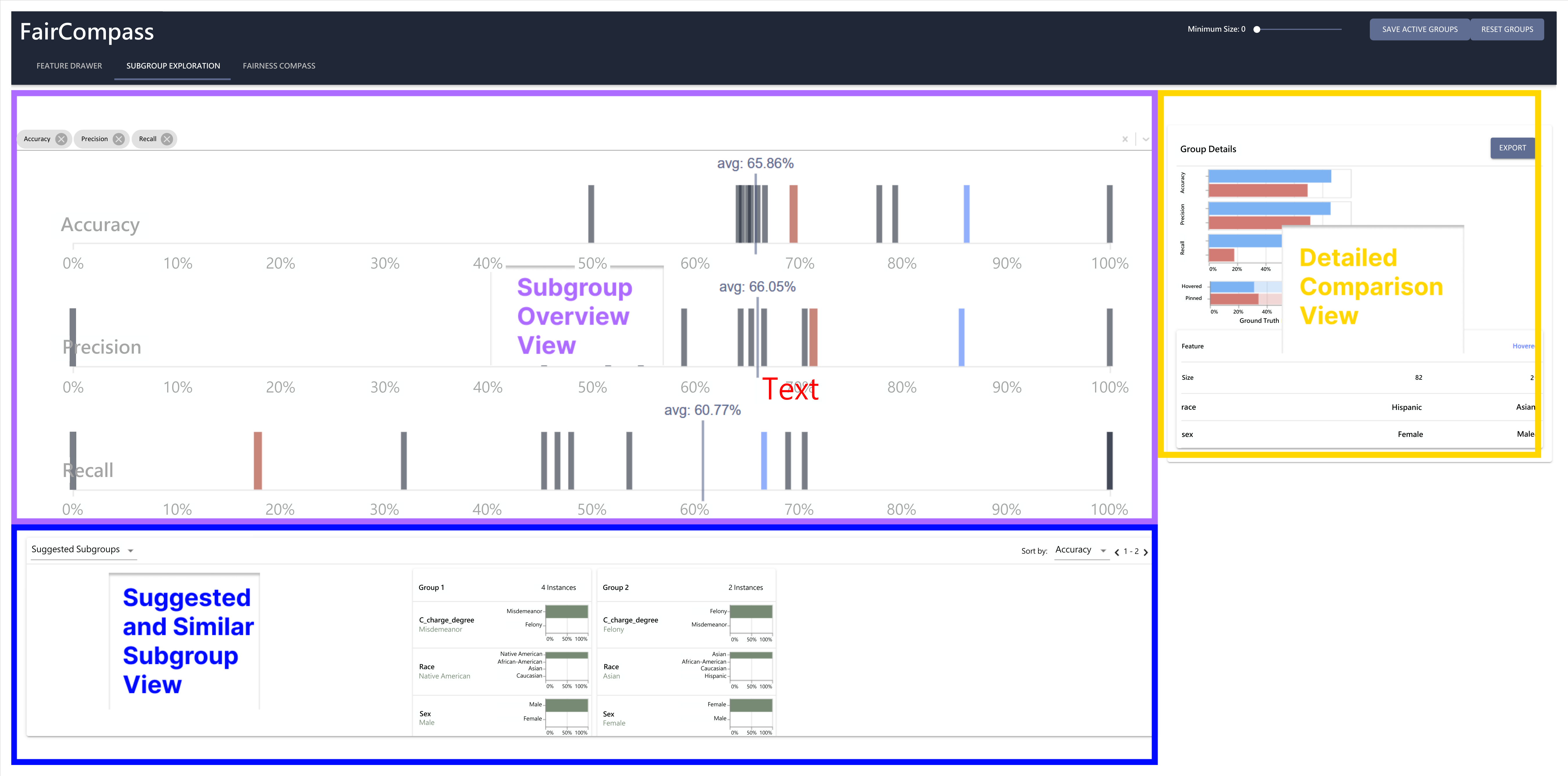}
\caption{The Subgroup Exploration Tab, which includes Subgroup Overview View, Suggested and Similar Subgroup View, and Detailed Comparison View} \label{fig3}
\end{figure*}
\subsection{Implementation}
In FairCompass, we have specified two primary views, which are the Subgroup Exploration Tab, and the Fairness Compass Tab. The Feature Distribution View is designed for individual interaction while simultaneously showing either Subgroup Exploration Tab or Fairness Compass Tab. The state of generated groups is persistent over the two different tabs, allowing users to switch between tabs to perform analysis on the same subgroups. The tab system allows any new information learned from the Fairness Compass tab, to be immediately explored or actioned through the Subgroup Exploration tab. 

To better support practitioners, split panes with movable dividers are used so that users can manipulate the size of each component in the tabs. This allows users to customise their viewing experience and prevent GUI clutter. Tooltips on hover for each button and tab are added to give users guidance and explain the functionalities.

\textbf{Feature Distribution View} is implemented in the form of a collapsible sidebar in FairCompass, as illustrated in Fig.~\ref{fig2}. We explicitly design the view collapsible to prevent clutter, as it is the panel that the user will spend the least time on, therefore the option is given to completely collapse it and provide more space for the tab contents. The functionality of this component is for viewing feature distributions and generating subgroups. Additionally, an additional active group saving functionality is implemented. To help users organise their thought processes, users can save a set of subgroups of interest for future revisit. Upon clicking on a saved set of subgroups in the saved subgroups list, the set is re-added to the currently active groups in the Subgroup Exploration and Fairness Compass tabs. This takes the cognitive stress off practitioners with ad-hoc analysis needs when there are many branches of information that need to be explored to gather evidence for a hypothesis.

\textbf{Subgroup Exploration Tab} in Fig.~\ref{fig3} includes the Subgroup Overview, Suggested and Similar Subgroups View and the Detailed Comparison View.

\begin{figure*}
\centering
\includegraphics[width=.75\textwidth]{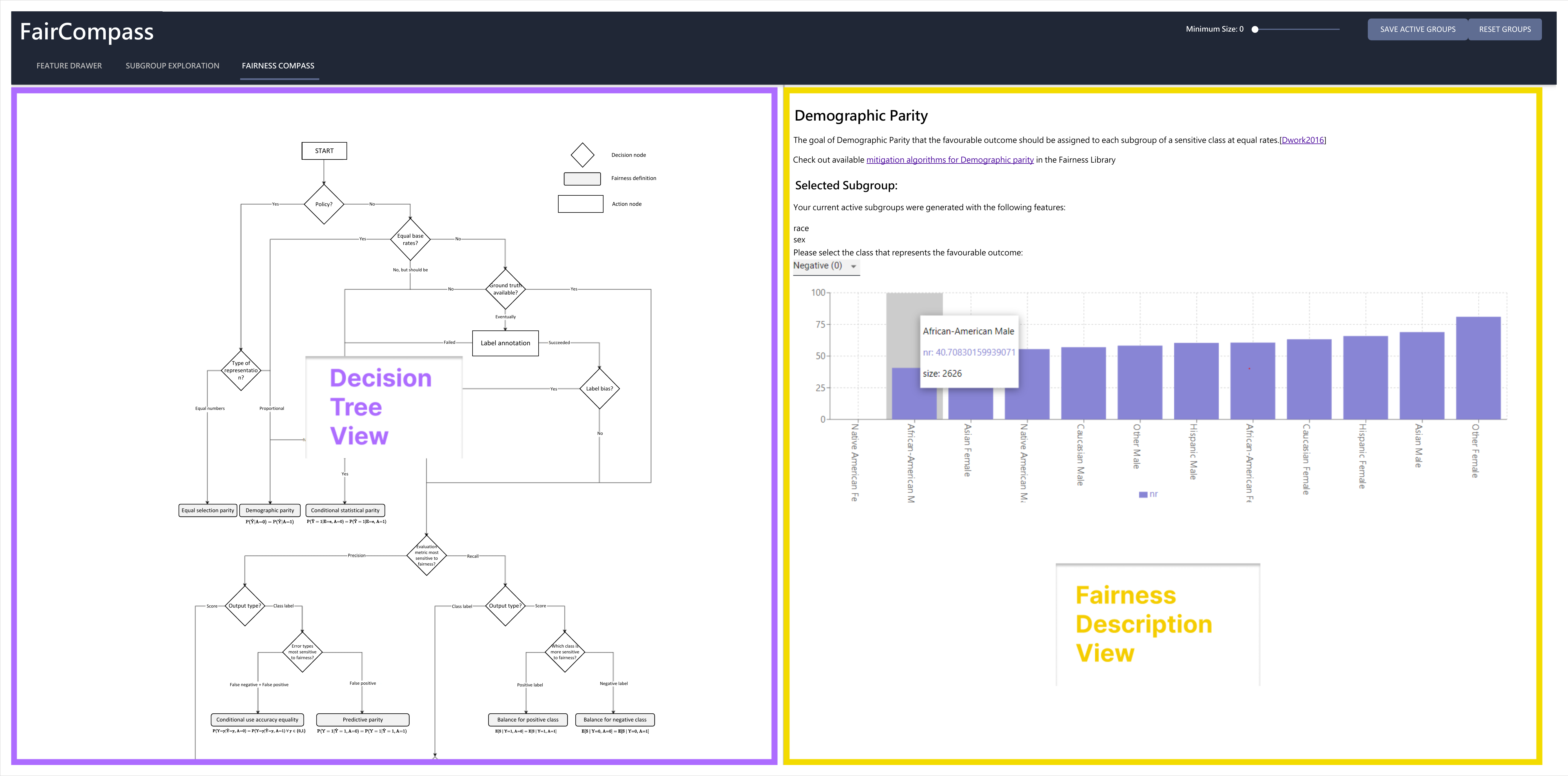}
\caption{The Fairness Compass Tab, which includes the newly added Decision Tree View using the Fairness Compass, and the newly added Fairness Description View.} \label{fig4}
\end{figure*}

\textbf{Fairness Compass Tab} consists of two panels, as shown in Fig.~\ref{fig4}. The Decision Tree View on the left shows a Fairness Compass decision tree~\cite{Ruf2021}. Missing explanations and formulas of the Fairness Compass are added using the information from~\cite{Ruf2021}, with minor adjustments to fit the context of the FairCompass application. Upon clicking on each node in the decision tree, descriptions of each node are shown in the Fairness Description View on the right panel. If the selected node is a fairness metric, visualisations are shown for the active subgroups, to assist users in assessing the current set of subgroups with the fairness metric. We have implemented bar plots and scatter plots, providing a more multifaceted presentation of the data. The visualisations chose to take the binary subgroup approach from the Fairness Compass and convert them into intersectional comparisons between multiple subgroups. With certain fairness definitions, users can customise the visualisations to align with their problem and application context. Consider a scenario where a user is assessing the active subgroups with the Demographic Parity definition. Demographic Parity is achieved when there is an equal proportion of favourable outcomes for all subgroups. The visualisation component for the definition has a dropdown that allows users to select the class in their dataset that represents the favourable outcome. If the user’s dataset is about providing opportunities, such as scholarship grant decisions for students, the user would choose the positive class as the favourable outcome. Conversely, if the user’s dataset is about recidivism prediction such as the COMPAS dataset, the user would select the negative class (no recidivism) as the favourable outcome. 
\subsection{Exploration, Guidance, and Informed Analysis Loop}
\begin{figure}
\centering
\includegraphics[width=\linewidth]{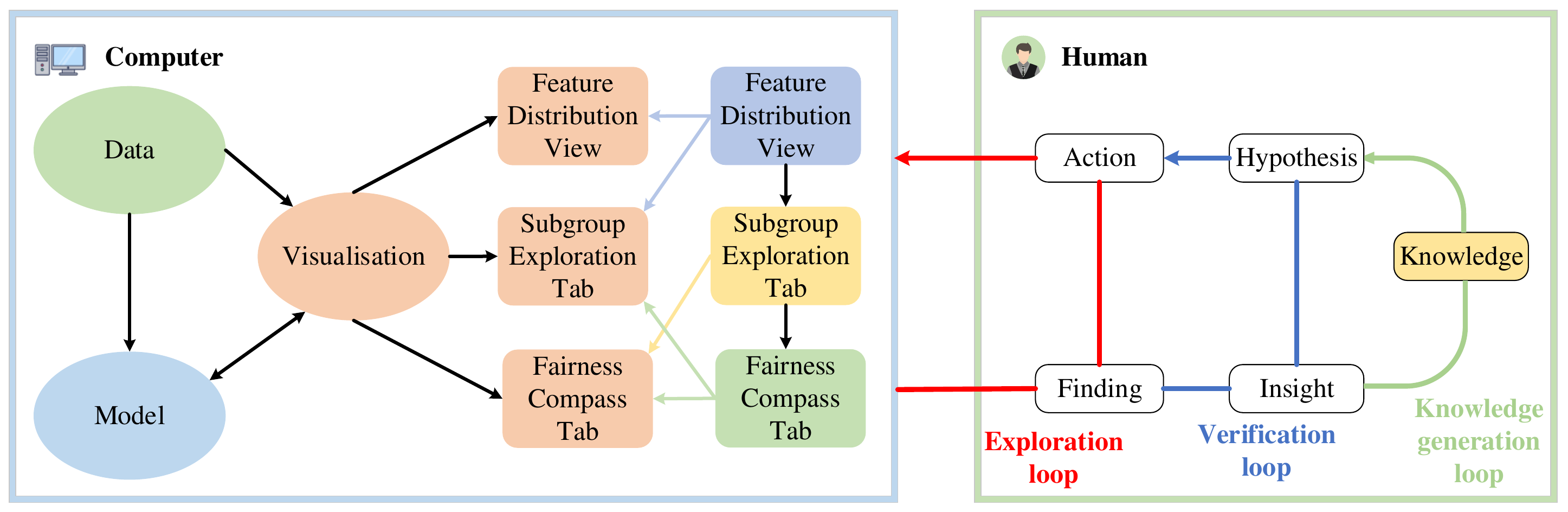}
\caption{The Knowledge Generation Model for Visual Analytics, extended with the newly proposed Exploration, Guidance and Informed Analysis loop, with details on how it works with the FairCompass components.} \label{fig5}
\end{figure}

We propose the Exploration, Guidance, and Informed Analysis conceptual framework, to provide a smoother integration between FairCompass and the Knowledge Generation Model in Fig.~\ref{fig5}. The design is made to support the implementation of this framework. While this section outlines how this framework supports the analysis process within the new design, the design of this framework intends to be generic across similar work and can be adjusted accordingly for other domains.

Exploration stage of the framework consists of the Feature Distribution View and Subgroup Exploration Tab of the new design. This stage is used to explore a high level overview of subgroups. The user builds a hypothesis in this stage and sets up a general objective for the rest of the analytical process. Guidance stage of the framework consists of the Feature Distribution View and Fairness Compass Tab of the new design. This stage is used to let users interact with the Fairness Compass and the different definitions available. Moving through the decision tree presents opportunities for users to discover more about the data and contextual factors surrounding it. Users can generate subgroups that they are interested in, test out different fairness definitions, and gather insight through the visualisations provided. In this stage, the user gathers more information to be able to action their objectives from the overview stage and produce insights as they move through each node in the decision tree. This gives practitioners with limited fairness knowledge adequate guidance to navigate through the task of selecting an appropriate fairness definition specific to the application context and the dataset. More ``trustworthy'' insights can be gathered from this process through the practitioner’s questioning, understanding, and reasoning using the fairness compass. Informed Analysis stage of the framework consists of both the Subgroup Exploration Tab and the Fairness Compass Tab. This stage occurs after a user has gained a good overall understanding of their dataset and enough fairness-related knowledge to analyse their hypothesis and findings from the Exploration stage. Through this, users will either be able to solidify their insights into new knowledge to begin a new iteration of the loop, or branch off into the investigation of related areas to support their central hypothesis.

\section{Evaluation on an Income Prediction System}
This section provides a use case and a walkthrough of how FairCompass can be used in fairness auditing. In this scenario, a machine learning practitioner is assigned by a financial institution to determine if an income prediction model is fair. The institution is planning on using this model in the development of an application that determines the loan size they are willing to grant an individual. The model classifies instances into positive (1) where the individual makes less than or equal to \$50,000 a year, or negative (0) where the individual makes more than \$50,000 a year. In this work, we use an effective income prediction model, which is a simple two-layer neural network with the Adult Income dataset from UCI Machine Learning Repository~\cite{UCI_data}. The practitioner wants to find out if the model perpetuates bias against disadvantaged individuals.
\begin{figure*}[htbp]
\centering
\includegraphics[width=.8\textwidth]{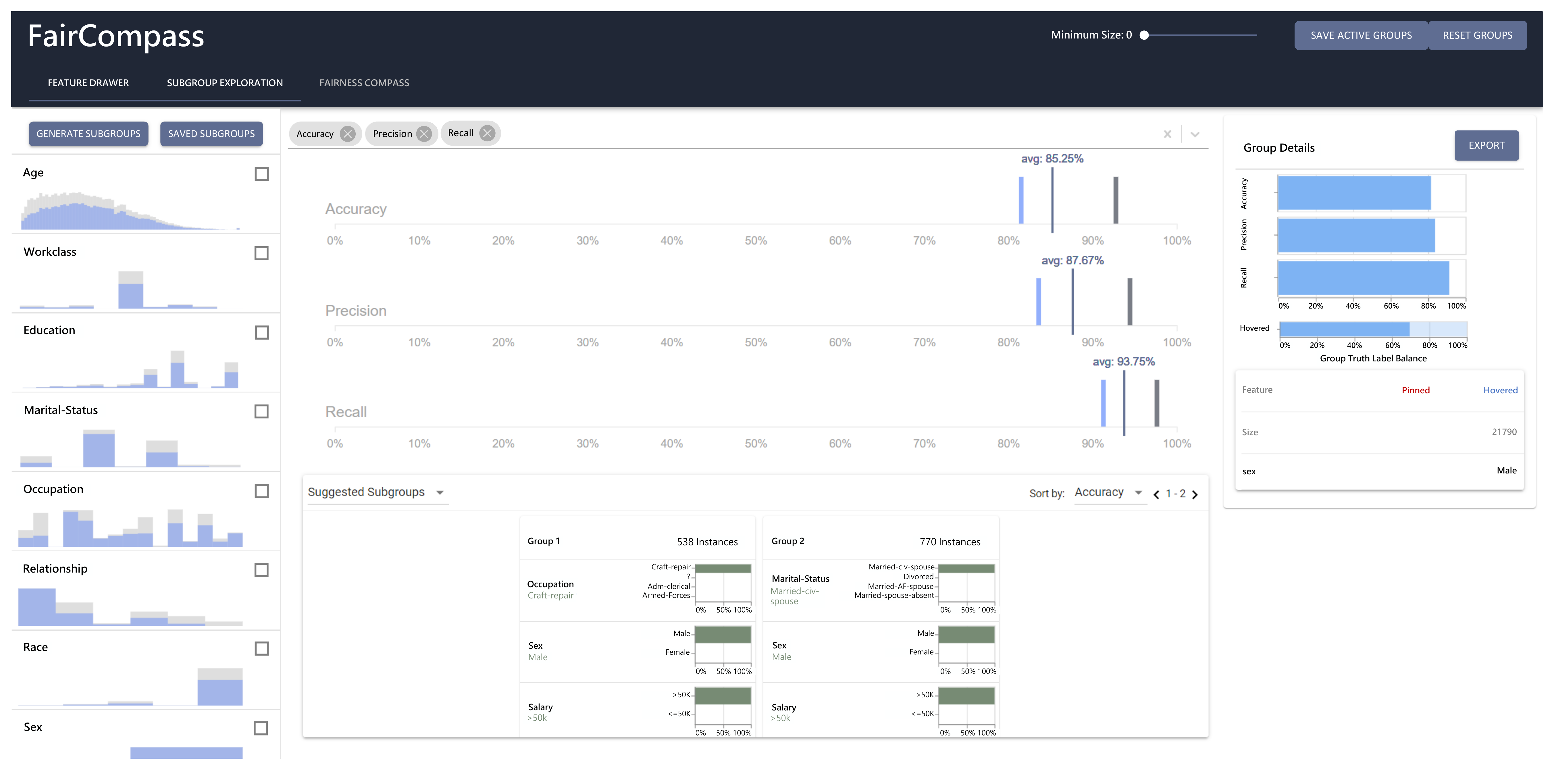}
\caption{A view of the Feature Distribution View and the Subgroup Exploration Tab. The accuracy, precision, and recall of the Male and Female subgroups are shown.} \label{fig6}
\end{figure*}
\paragraph{Iteration 1: Exploration} She first looks at the feature distribution of the data and finds that the sample size of males is twice the size of females. She decides to generate subgroups to explore based on gender. Upon generation of the subgroups, she sees that the female subgroup has higher accuracy, and the male subgroup has higher precision and recall. She wonders if there are more complex relationships present in the prediction results of these subgroups and would like to explore this more but does not know where to begin. This is shown in Fig.~\ref{fig6}.

\begin{figure}
\centering
\includegraphics[width=0.8\linewidth]{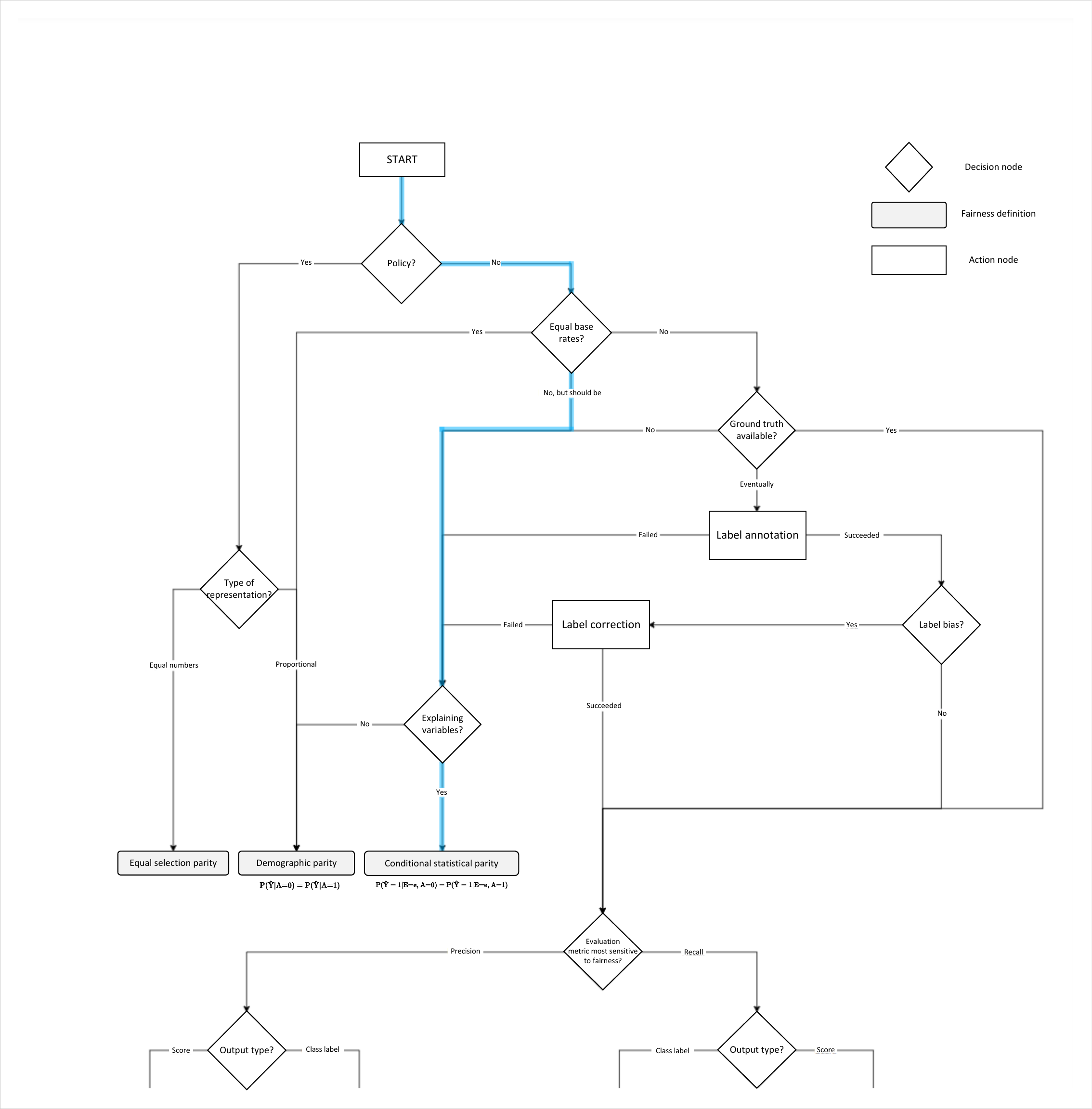}
\caption{The path that the practitioner undertakes in the FairCompass.} \label{fig7}
\end{figure}

\paragraph{Iteration 1: Guidance} She clicks on the Fairness Compass Tab and navigates through the decision tree:
\begin{itemize}
  \item Policy: The practitioner confirms with her team and the policymakers in her organisation that there are no anti-discrimination policies in place for the purpose of home loans based on an individual’s income, and the organisation is not planning on implementing affirmative action at this stage. However, there are laws that prohibit lending bias against individuals based on race, sex, marital status, and age. Although these policies do not apply to granting home loans on the basis of income, the practitioner thinks that it is important to investigate the sensitive attributes in lending bias for the purpose of her task. She decides to explore the sex attribute first but notes down the others for further exploration. She continues down the \textit{No} option.
  \item Equal base rates: From the high-level overview of the data that the practitioner has performed, she knows that 2/3 of all instances in the data belong to the male group. She questions if this dataset is appropriate for the application context, due to the disparity in sample size. According to the explanation provided in the Fairness Description View, assuming equal base rates can benefit historically discriminated groups in decisions related to opportunity. She continues down the \textit{No, but should be} option.
  \item Explaining variables: The practitioner looks through the Feature Distribution View again to see all attributes of the data. She finds that there are several explaining variables for outcome in this dataset. Features such as occupation and working hours are factors that can explain disparities in salary. She continues down the \textit{Yes} option, as presented in Fig.\ref{fig7}.
\end{itemize}

\begin{figure*}[htbp]
\centering
\includegraphics[width=.8\textwidth]{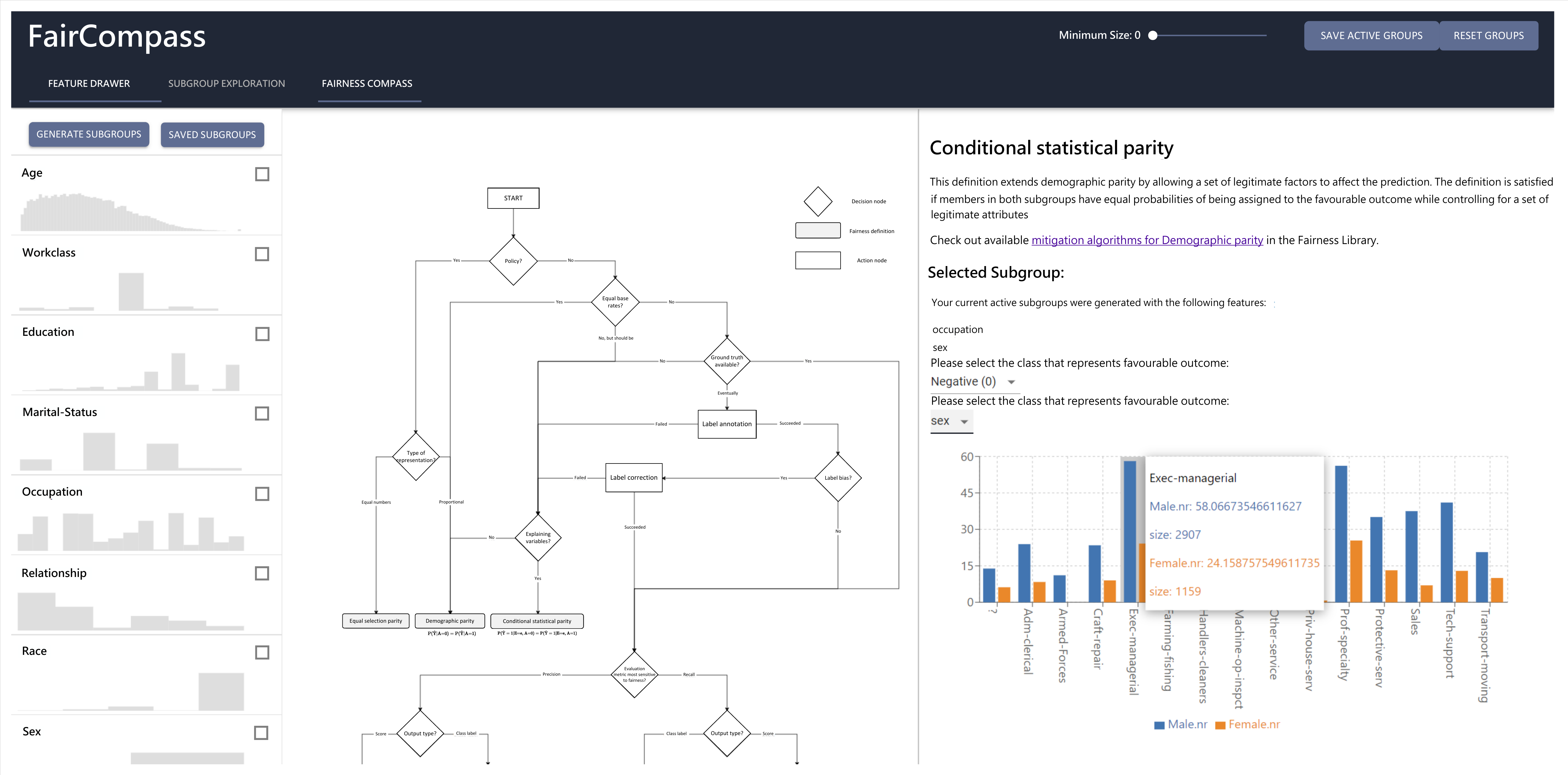}
\caption{The conditional statistical parity definition and bar chart visualisation for the sex, and occupation subgroups on the Fairness Description View of the Fairness Compass Tab.} \label{fig8}
\end{figure*}
The decision tree leads her to the Conditional statistical parity definition. This definition is an extension of Demographic parity with consideration of a set of legitimate attributes (explaining variables) that affect income. The practitioner wants to ensure that there are fair outcomes between male and female individuals, applying this fairness definition means that the model should assign equal proportions of favourable outcomes to male and female candidates of the same working hours, occupation, etc. The practitioner decides to explore the relationship between sex and each explaining variable separately. 

Shown in Fig.~\ref{fig8}, she first generates a new set of subgroups based on sex and occupation and clicks on the Conditional statistical parity node. Two dropdowns appear, the first asking her to select the favourable outcome, and the second asking her to select the sensitive attribute. She selects the negative class ($<$ 50k) as the favourable outcome, as larger loans are granted to those of higher income in this context, and selects sex as the sensitive attribute. According to the bar chart, she finds that for all occupations, female candidates have a lower negative rate, meaning that for each occupation, the model predicted proportionally fewer female candidates to have incomes higher than 50k, in comparison to male candidates. Upon this discovery, the practitioner decides to take a closer look at the Male Exec-managerial subgroup and Female Exec-managerial subgroup, as these subgroups have a proportionally large sample size and a large discrepancy of 34\% in negative rate according to the bar chart visualisation.

\begin{figure*}
\centering
\includegraphics[width=.8\textwidth]{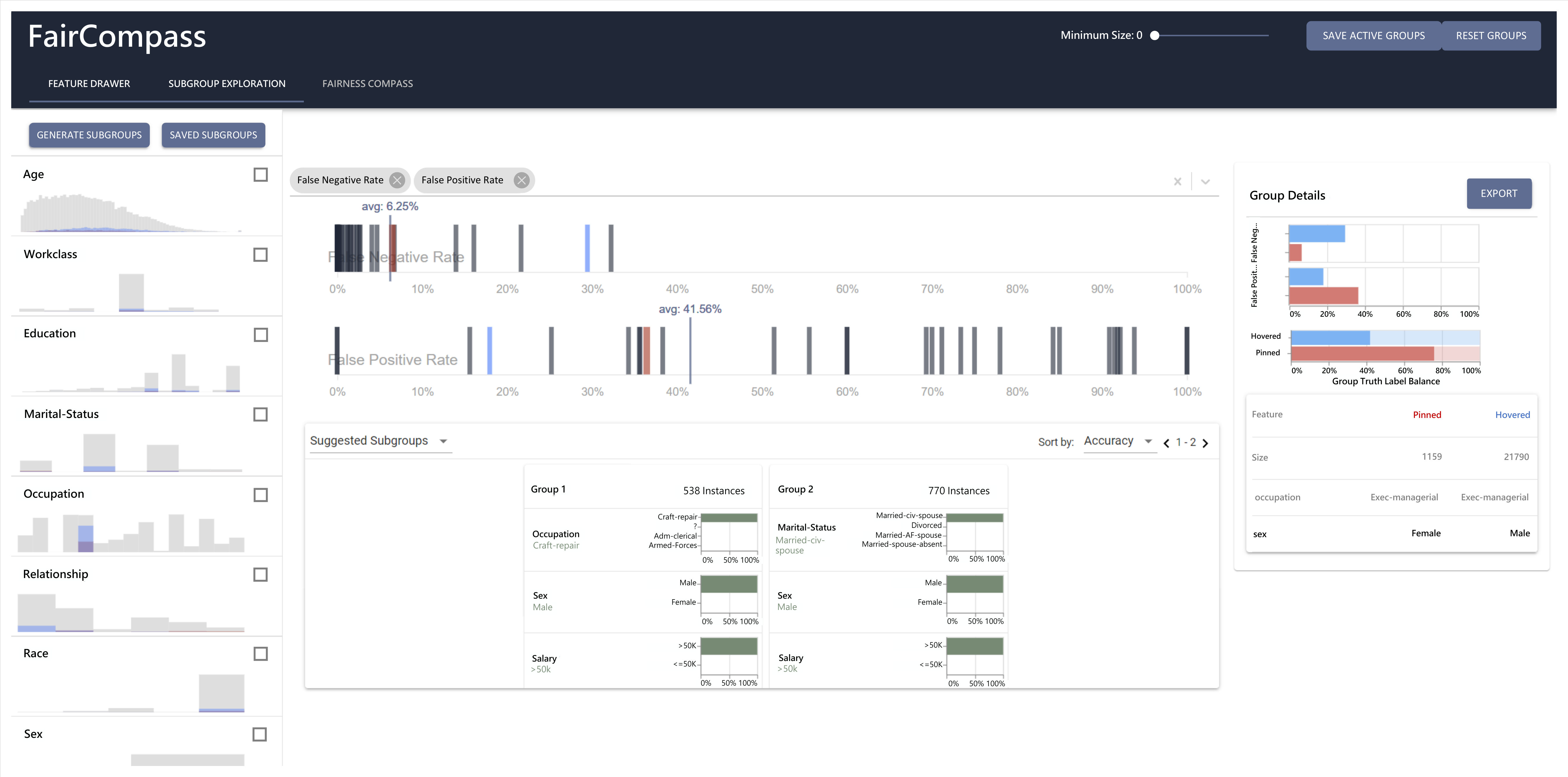}
\caption{The false negative rate and false positive rate of the sex, and occupation subgroups in the Subgroup Exploration Tab.} \label{fig9}
\end{figure*}

\paragraph{Iteration 1: Informed Analysis} Informed by her new insight into the negative rate discrepancies she returns to the Subgroup Exploration Tab and selects the false negative rate and false positive rate metrics for further investigation. She pins the Male Exec-managerial subgroup, and hovers over the Female Exec-managerial subgroup, to view a comparison of the metrics in the Detailed Comparison View. In the figure below, the Detailed Comparison View shows the comparison of these metrics for the Exec-managerial Male and Exec-managerial Female subgroups. She finds that the Female Exec-managerial manager group has an almost 25\% lower false negative rate and an almost 20\% higher false positive rate. This means that instances in the Male Exec-managerial subgroup were more likely to be falsely classified as having an income higher than 50k, and less likely to be falsely classified as having an income less than or equal to 50k, compared to their female counterparts. The practitioner has now gained more “trustworthy” evidence of unfairness between the female and male subgroups, however, she wants to gather more evidence surrounding the unfair treatment of female individuals in this model before drawing a conclusion. This is shown in Fig.~\ref{fig9}.

\begin{figure*}
\centering
\includegraphics[width=.8\textwidth]{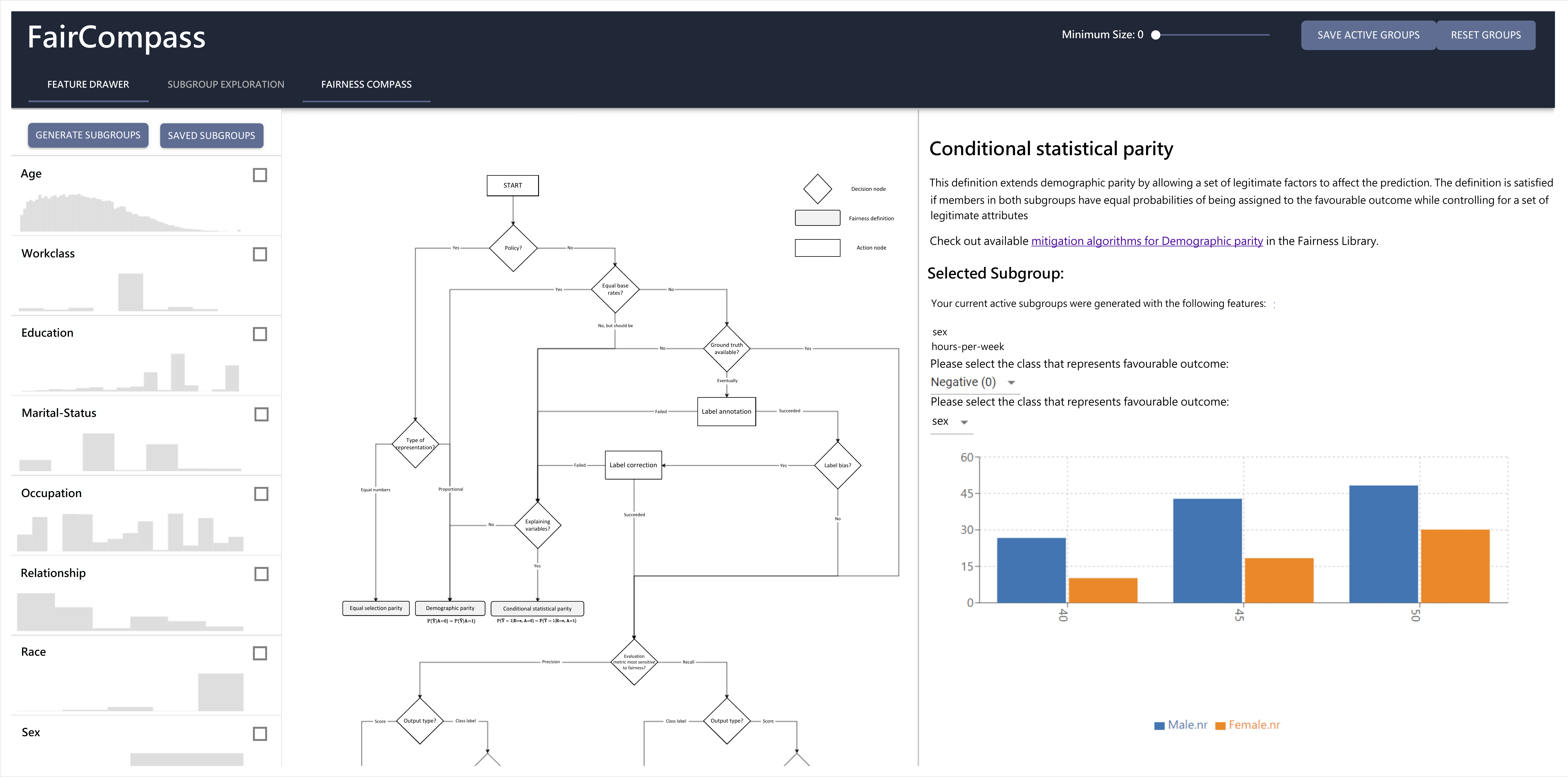}
\caption{The conditional statistical parity definition and bar chart visualisation for the sex, 40, 45, 50 working hours subgroups on the Fairness Description View of the Fairness Compass Tab.} \label{fig10}
\end{figure*}

\paragraph{Iteration 2: Exploration} As she has previously identified hours of work as an explaining variable assisted by the Fairness Compass, she decides to compare the prediction results between male and female individuals that work the same number of hours. Upon viewing the feature distribution for hours of work in Fig.~\ref{fig10}, she finds that almost half of all instances are of 40 hours. Due to the drastic difference in sample size, the practitioner decides that she will select the top three values with the largest sample size to avoid analysing data with representation bias. The practitioner selects 40, 45, and 50 for the hours of work, along with the sex attribute and generates subgroups. She finds that the Male 50 and Male 45 subgroups have the lowest accuracy, precision and recall of all groups, and the Female 40 Subgroup has the highest accuracy, precision and recall. Although these values suggest that the Female subgroup has better performance using these metrics, the practitioner would like to take a closer look at the visualisation on the Fairness Compass Tab.

\paragraph{Iteration 2: Informed Analysis} Upon checking if the subgroups achieve the conditional statistical parity definition in the Fairness Compass Tab, she finds that the negative rate for male subgroups is higher than for female subgroups by more than 15\% for all three working hours. To further explore this phenomenon, she goes back to the Subgroup Exploration Tab to view the false negative rate and false positive rate of these groups. She finds that all female subgroups have a below average false negative rate, and all male subgroups have an above average false negative rate. The converse is true for the false positive rate metric, further solidifying the bias against female subgroups. After checking with two explaining variables, the practitioner is now more confident that this model has a tendency to classify female individuals as having a lower income compared to their male counterparts. The practitioner has concluded that this model is unfair and should not be used in the application of granting home loans. She decides to further explore the other sensitive attributes mentioned in the lending bias policies she discovered earlier, to try to find other sources of discrimination, before submitting a formal report of her findings to her boss.

This use case presents a practical usage of the system, where the practitioner completes two iterations of the Exploration, Guidance and Informed Analysis loop. Through this analysis process, she also engages the exploration loop, verification loop, and knowledge generation loop of the Knowledge Generation Model for Visual Analytics. The overall evaluation of the use case represents an effective outcome of the system. However, we would like to highlight that, the analysis process can vary and diverge from our example depending on the problem context, dataset, and the practitioner's personal preferences in performing analysis. 

\section{Limitations and Future Work}
In this section, we highlight the limitations of FairCompass, and discuss recommendations for future work in this direction. 

\subsubsection{Polishing the Guidance stage and Extending the Fairness Compass} Ruf \& Detyniecki~\cite{Ruf2021} conclude that there is no silver bullet to overcoming AI bias and that the Fairness Compass is not the last word on the subject, as fairness research is constantly advancing. The guidance stage offered in our design is rather limited in its coverage of the complex landscape of fairness research. Therefore, more work can be done to develop a comprehensive taxonomy of biases and harms, as well as implement individual and subgroup notions of fairness in the Fairness Compass, in a way that is easily digestible by practitioners who are unfamiliar with the topic. 

\subsubsection{Biases in Human-in-the-loop} Algorithm to use biases may be introduced in the use of the application and the analytical process. Holstein et al.~\cite{Holstein2019} note that several participants in their study emphasised the importance of taking into consideration the human biases embedded at each stage of the machine learning life cycle. An example is presentation bias, which is a form of bias that stems from how the information is presented to users, leading them to come to biased conclusions~\cite{BaezaYates2018}. With our suggestion of human-in-the-loop fairness auditing, there needs to be an increased awareness of mitigating human biases. While the Fairness Compass helps to relieve the burden of fairness placed on practitioners, an overreliance on it or the belief that conclusions made from the tree are always the "correct" solution, can be detrimental to practitioners’ pursuit of fairness. This presents a case of Sandvig’s neutrality fallacy~\cite{Sandvig2014}, where users believe that the system is correct even in situations where it is obvious that it is not. The aim of the Guidance stage of our proposed framework is not to teach users everything they need to know about fairness, but to introduce them to concepts that they may be unfamiliar with and encourage them to seek out further resources if needed. Future work for mitigating use bias can include the creation of guidelines for preventing use bias and enforcement of implicit bias training programs specific to practitioners who are using interactive systems for fairness auditing.

\subsubsection{Developing Domain Specific Tools} Both FairVis and the Fairness Compass are designed for very general use of fairness auditing. This means that they may be insufficient in solving real-world problems that require more domain specific tools. FairVis currently only supports binary classifiers, which means that users are unable to audit the fairness of other output types such as those of ranking algorithms, recommendation systems, voice and facial recognition systems etc. The k-means clustering method of generating suggested subgroups may also not be able to cover domain specific situations where features have different weightings of importance and have complex relationships with one another. In these instances, FairVis’s subgroup recommendation functionality may obscure information essential to decision making. The Fairness Compass decision tree may be inapplicable to specific application contexts. For example, we have found that when using the tree in the context of utilising the COMPAS dataset for recidivism prediction, it is reasonable to end on the demographic parity definition. Dwork et al.~\cite{Dwork2012} argued that demographic parity is a flawed definition in practice by proving three scenarios where the definition is maintained yet produces a blatantly unfair result from an individual fairness perspective. Minority communities are controlled and policed more frequently~\cite{Suresh2021}, resulting in a self-fulfilling loop of higher arrests and higher base rates. Using the demographic parity definition may further perpetuate pre-existing biases against these individuals in the data, as the definition requires groups to be assessed the same when they have historically not been treated the same. For future work, the general approach proposed in this paper can be adjusted for specific domain, and take into account real life problems where fairness is heavily context dependent.


\subsubsection{Enforcing Fairness at a Higher Level} This project aims to give practitioners better support in making fairness-related decisions, due to the fact that the responsibility of ensuring the fairness of a system is often disproportionally assigned to them when this should not be the case. It is ultimately the organisation’s responsibility to ensure that there are fairness measures in place, or that practitioners receive an adequate amount of support to be able to carry out these procedures. The responsibility of AI fairness within an organisation extends to adjacent or higher-level actors such as data scientists, domain experts, policymakers and stakeholders. While our project lays the groundwork for operationalising fairness, there is a need for organisations and institutions to prioritise fairness and install formal processes that enforce fairness, and eventually cement these practices as standard procedures in the industry.

\section{Conclusion}
In this paper, we firstly review the existing fairness tools and identify areas of challenges suggested by fairness experts and practitioners. The major issues often brought forward include the identified challenges of overemphasis on technical solutions and difficulties in operationalising fairness. To address these concerns, we suggest a novel approach by leveraging both technical and non-technical approach integrated within a visual analytics system, to streamline the bias auditing process with a human-centred design. We demonstrate our proposed system of FairCompass. We also propose to incorporate the Exploration, Guidance, and Informed Analysis loop, in order to apply the Knowledge Generation Model for Visual Analytics to FairCompass, and also take a first step in structuring the novel approach for future works. We anticipate FairCompass to steer more research for the intersection between visual analytics and fairness in machine learning, encouraging experts to prioritise operationalising fairness in practice.

\balance
\bibliography{mybibliography}
\bibliographystyle{ieeetr}


\end{document}